\documentclass[journal]{IEEEtran}
\usepackage{cite}
\usepackage{amsmath,amssymb,amsfonts}
\usepackage{algorithmic}
\usepackage{graphicx}
\usepackage{textcomp,booktabs}
\usepackage{xcolor}
\usepackage{multirow}
\usepackage{lipsum}
\usepackage{float}
\def\BibTeX{{\rm B\kern-.05em{\sc i\kern-.025em b}\kern-.08em
    T\kern-.1667em\lower.7ex\hbox{E}\kern-.125emX}}

\begin{document}
\title{Self-Supervision and Spatial-Sequential Attention Based Loss for Multi-Person Pose Estimation}

\author{Haiyang Liu,~\IEEEmembership{Student Member,~IEEE,} 
        Dingli Luo,
        Songlin Du,~\IEEEmembership{Member,~IEEE,}
        and~Takeshi~Ikenaga,~\IEEEmembership{Senior Member,~IEEE}
\thanks{H. Liu, D. Luo and T. Ikenaga are with the Graduate School of Information,Production and Systems, Waseda University, Kitakyushu-shi, 8080135 Japan. (e-mail: haiyangliu@toki.waseda.jp, luodingli@toki.waseda.jp and
ikenaga@waseda.jp).}
\thanks{S. Du is with the School of Automation, Southeast University, Nanjing, 210096 China. (e-mail: sdu@seu.edu.cn)}}%

\markboth{Journal of \LaTeX\ Class Files,~Vol.~14, No.~8, August~2015}%
{Shell \MakeLowercase{\textit{et al.}}: Bare Demo of IEEEtran.cls for IEEE Journals}

\maketitle

\begin{abstract}
Bottom-up based multi-person pose estimation approaches use heatmaps with auxiliary predictions to estimate joints positions and belonging at one time. Recently, various combinations between auxiliary predictions and heatmaps have been proposed for higher performance, these predictions are supervised by the corresponding L2 loss function directly. However, the lack of more explicit supervision results in low features utilization and contradictions between predictions in one model. 
To solve these problems, this paper proposes (i) a new loss organization method which uses self-supervised heatmaps to reduce prediction contradictions and spatial-sequential attention to enhance networks' features extraction; (ii) a new combination of predictions composed by heatmaps, Part Affinity Fields (PAFs) and our block-inside offsets to fix pixel-level joints positions and further demonstrates the effectiveness of proposed loss function. 
Experiments are conducted on the MS COCO keypoint dataset and adopting OpenPose as the baseline model. Our method outperforms the baseline overall. On the COCO verification dataset, the mAP of OpenPose trained with our proposals outperforms the OpenPose baseline over 5.5\%.
\end{abstract}

\IEEEpeerreviewmaketitle

\section{Introduction}
\IEEEPARstart{M}{ulti-Person} pose estimation refers to estimating pixel-level joints positions for each person in images. Specifically, the number of people is larger than one in each image, which makes judging joints belonging and locating joints positions to become two necessary tasks. 
Multi-Person estimation has received widespread attention in plenty of applications, such as sports analysis\cite{zhu2007human}, human-computer interaction\cite{cai2015effective}, and video surveillance\cite{gualdi2008video}. 
With the development of Convolutional Neural Networks (CNNs), most of recent pose estimation frameworks are based on CNNs\cite{huang2017coarse,carreira2016human,chu2017multi,lifshitz2016human,belagiannis2017recurrent,ramakrishna2014pose} and achieve superior performance. These methods to generate multi-person pose could be divided into two types: \textit{Top-Down} methods\cite{newell2016stacked,he2017mask,fang2017rmpe,chen2018cascaded,papandreou2017towards,xiao2018simple,li2019crowdpose,li2019rethinking,zhang2019human,chou2018self} and \textit{Bottom-Up} methods\cite{cao2017realtime,cao2018openpose,newell2017associative,kocabas2018multiposenet,papandreou2018personlab,nie2019single,kreiss2019pifpaf,varadarajan2018greedy,gkioxari2013articulated, liu2020resolution}. The difference between these two kinds of methods is whether the time consumption increases linearly with the number of people in the image. \textit{Top-Down} methods will generate human boundary boxes\cite{fang2017rmpe} or proposals\cite{he2017mask} in the first step and then estimate joints positions for each proposal, the processing speed is related to the number of boundary boxes or proposals in one image. For the \textit{Bottom-Up} methods, except estimating joints positions predictions, the model will also generate one or more auxiliary predictions for assembling joints, these procedures will be finished by only look at the image once. In this paper, we focus on the \textit{Bottom-Up} methods, which are commonly used in real-time applications\cite{angelini20192d} because of its speed advantages.  

\begin{figure}
  \includegraphics[width=\linewidth]{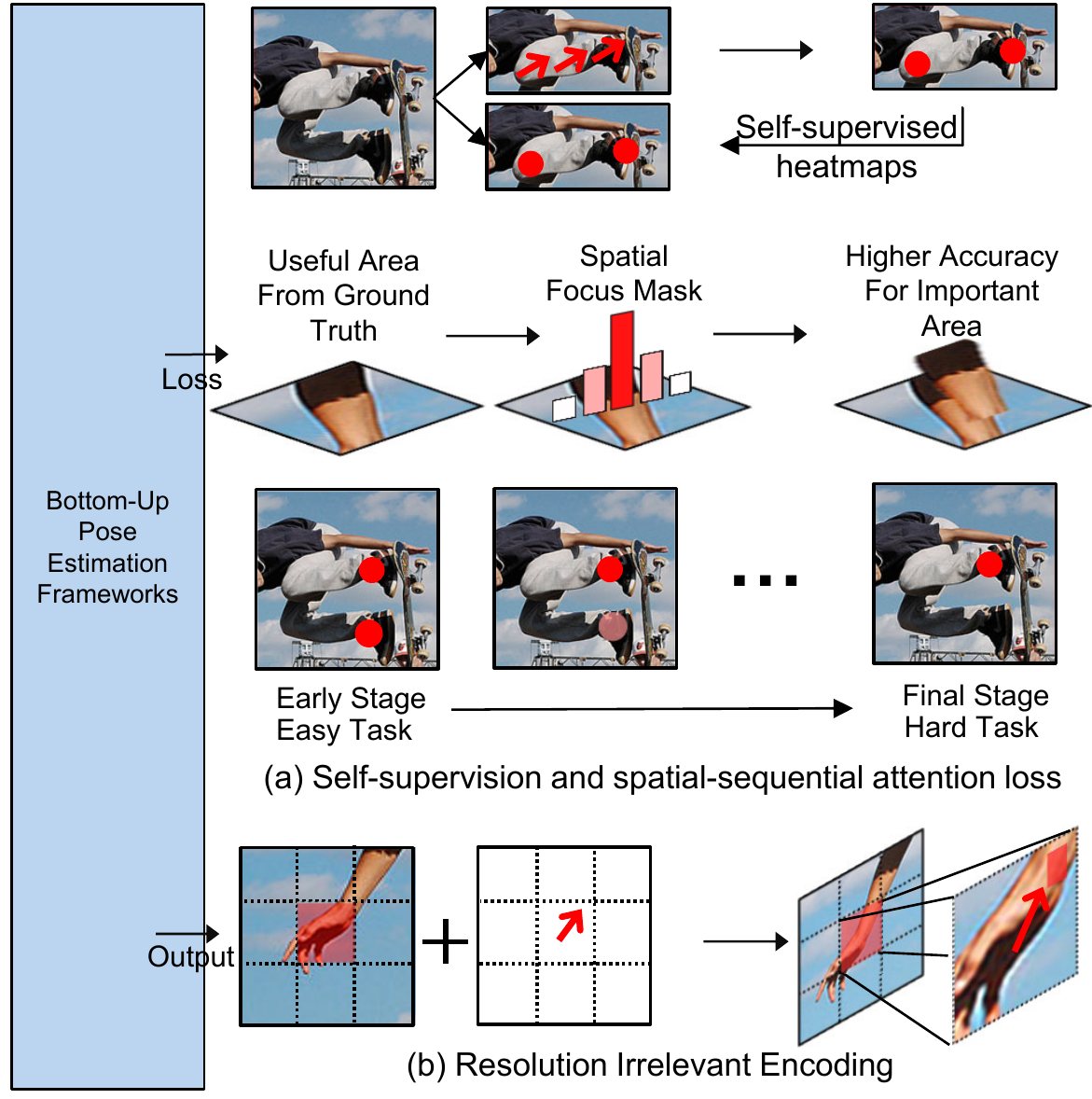}
  \caption{Our network-agnostic supervision. (a) Self-Supervision and Spatial-Sequential Attention Loss to make the most of pose networks performance. (b) Resolution Irrelevant Encoding for improving the pixel-level joints' location accuracy.
}
  \label{fig:concept}
\end{figure}

However, the \textit{Bottom-Up} methods are difficult to realize as high accuracy as \textit{Top-Down} methods for the following reasons. Firstly, the addition of auxiliary predictions leads the indeterminate influence to the final accuracy. The Associative Embedding\cite{newell2017associative} has joints jitter error even use groundtruth joints positions as inputs, which means the auxiliary predictions they designed leading to negative influences to joints positions estimation. However, the Part Affinity fields (PAFs), have been proofed that it has a positive influence to locate joints in the OpenPose model\cite{cao2017realtime}. Secondly, different person size varies in one image dramatically. For example, the smallest person only occupies about 1/100 area of one image but the largest person could occupy the image with the upper limb. Lastly, the single person joints refinement is easier to implement compares with the multi-person case. For example, the Online Hard Example Mining (OHEM)\cite{chen2018cascaded} and Graph Neural Networks (GNNs)\cite{zhang2019human} have been applied for single person pose refinement, however, these methods are hard to implement in the multi-person case directly because of the uncertainty of joints numbers.

Much research has been investigated on the \textit{Bottom-Up} methods because of speed stability in real-world applications. The tradition method of \textit{Bottom-Up} based pose estimation is part-based\cite{hua2005learning,felzenszwalb2005pictorial,ramanan2005strike,andriluka2010monocular,andriluka2009pictorial,pishchulin2013poselet,yang2012articulated,johnson2010clustered}. They accumulate prior knowledge of 2D body shapes for their data-driven Monte Carlo algorithm. Since Deep Neural Networks (DNNs)\cite{hinton2012improving} have achieved superior performance in prior knowledge learning, it is expanded into various fields. Especially, the Conventional Neural Networks\cite{krizhevsky2012imagenet} have been shown to outperform DNNs on this task\cite{wei2016convolutional,ouyang2014multi,tompson2015efficient,tompson2014joint,yang2011articulated,toshev2014deeppose,fragkiadaki2015recurrent,bulat2016human,chu2017multi,yang2017learning,chen2017adversarial,tang2018deeply,ke2018multi} as CNNs can capture spatial information in shared weights simultaneously. At the same time, the MicroSort COCO keypoints detection dataset\cite{ruggero2017benchmarking} and the benchmark has been published, The OpenPose\cite{cao2017realtime}, which combined part-based methods and CNNs, archived first place in the COCO keypoints task in 2016. The network will generate Part Affinity Fields (PAFs) as auxiliary predictions for assembling joints. 
After 2017, the research of \textit{Bottom-Up} based pose estimation is main focus on proposing more efficient auxiliary predictions\cite{newell2017associative,papandreou2018personlab,nie2019single,kreiss2019pifpaf}. Associative Embedding\cite{newell2017associative} uses several equal resolution images as auxiliary predictions, the difference between pixels value in the joints region stands for the different person belonging. Recently, the single-stage multi-person pose machines\cite{nie2019single} proposes a new auxiliary prediction by embedding the limb length information into previous part-based predictions. The PifPaf\cite{kreiss2019pifpaf} proposes a combination of auxiliary predictions by combining G-RMI\cite{papandreou2017towards} proposed offsets and their modified middle-level offsets\cite{papandreou2018personlab}.

So far, lots of auxiliary predictions combinations' effectivity has been proofed. The simplest loss function, e.g., L2 loss or L1 loss, is commonly used in these works. These simple loss function selections are usually for controlling experimental variables but result in several problems.

\textbf{Contradiction between predictions}. Firstly, the relationship between auxiliary predictions and heatmaps (joints location predictions) has not been supervised to learn from the loss function level. The results of OpenPose demonstrate that there have contradictions between PAFmaps and heatmaps results, i.e., The heatmaps and PAFmaps results stand for 'person-exist' regions for one image. In some cases, the heatmaps results are correct while the PAFmaps results are incorrect. 

\textbf{Inefficient features utilization}. The post-processing algorithm, which uses the auxiliary predictions and heatmaps to generate the final pose result, is necessary for \textit{Bottom-Up} based pose estimation methods. But not all information (pixels value) of auxiliary predictions has been used in the post-processing algorithm\cite{cao2017realtime,cao2018openpose,wang2008multiple,newell2017associative,nie2019single,kreiss2019pifpaf}, which means different regions of auxiliary predictions have different importance but be treated the same way by simple loss functions. 

\textbf{Demerits for multi-stage loss}. Excepting generate several auxiliary predictions, most State-Of-The-Art (SOTA) pose networks use multi-stage structures\cite{cao2017realtime,newell2017associative,nie2019single} to refine predictions. The different stages are usually supervised by the same loss function, e.g., Mean Square Error (MSE) loss. The performance of OpenPose\cite{cao2017realtime} and Hourglass\cite{newell2016stacked} demonstrate that there is only about 0.6\% mAP improvement for the later network stages. i.e., stage 4 to 6 in the OpenPose model and stage 5 to 8 in the Hourglass model. Adopting the same loss function in a multi-stage network leads to the unbalanced learning difficulty for different stages, there are few new features for the final several stages to learn for refining the performance\cite{li2019rethinking}.

\textbf{Demerits for higher-level supervision}. In our opinion, both auxiliary predictions and loss functions belong to supervision strategies. The auxiliary predictions are high-level supervision that defines the network outputs' types; the loss functions are lower-level supervision that defines how does the network generate expectant outputs\cite{lin2017focal}. For the high-level supervision, the resolution of inputs and outputs predictions in the previous works\cite{cao2017realtime,kreiss2019pifpaf} is different because of the down-sample layers in the network. There has an inevitable up-sample error for recovering the predictions to the resolution of inputs, which is a negative impact on the model performance. 

In summary, to solve the aforementioned challenges for Bottom-up based pose estimation methods, this paper proposes the self-supervision and spatial-sequential attention based loss function for the three problems in lower-level supervision. Besides, we make a new combination of predictions composed of heatmaps, Part Affinity Fields (PAFs), and our proposed block-inside offsets, which is called resolution irrelevant encoding, to fix pixel-level joints positions. The new combination is also for further demonstrating the effectiveness of our proposed loss functions. These proposals are network-agnostic and could be implemented for common \textit{Bottom-Up}\cite{cao2017realtime,papandreou2018personlab,kocabas2018multiposenet,kreiss2019pifpaf,wang2018magnify} frameworks. 

\begin{figure*}
  \includegraphics[width=\linewidth]{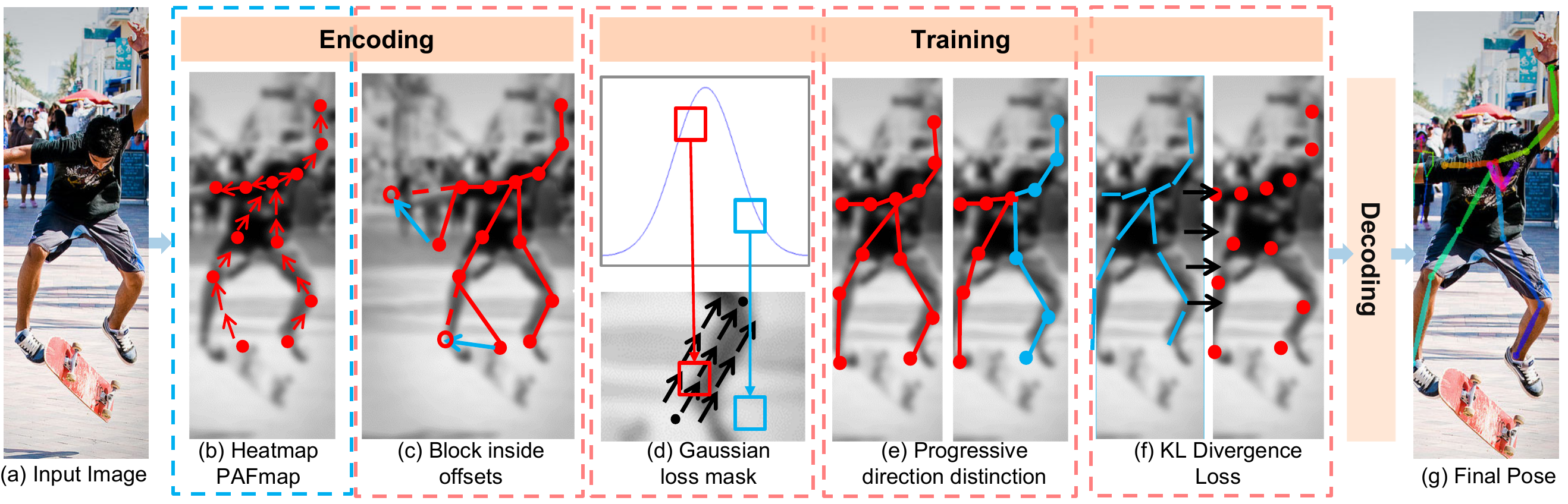}
  \caption{The system architecture of our methods. The (a) input image will be encoded into (b) heatmap, Part Affinity Fields map (PAFmap), and our (c) block inside offsets. The L2 loss function to supervise the multi-stage CNN will be reorganized by Spatial-Sequential Attention ((d) and (e)) and Self-Supervised Heatmap (f). Finally, the pose results (g) will be generated by the decoding algorithm.}
  \label{fig:framework}
\end{figure*}

Overall, as shown in Fig. \ref{fig:concept}, this paper's contributions are two folds:
\subsubsection{New loss functions} we design a new organization method containing two-part: Self-Supervision loss for explicit learning the relationship between auxiliary predictions and heatmaps. The self-supervised heatmaps loss calculation method is proposed to reduce the contradiction between predictions. Spatial-Sequential Attention adopts Gaussian distribution based loss weights for different pixels' loss calculation in auxiliary predictions, and adopts progressive punishment for different stages to guide the network focusing on valuable pixel-level information and gradual distinguishing joints left-and-right directions, respectively. 
\subsubsection{A new encoding method} Resolution Irrelevant Encoding (RIE) is proposed as additional high-level loss function. The block inside offsets regression task is added for guiding the network to estimate joints' precise positions into two steps.

Furthermore, there is no clear conclusion that which one is the best because of the difference in the backbone network chosen for previous works. e.g. the OpenPose\cite{cao2017realtime} uses VGG\cite{simonyan2014very}, the Single-Stage Multi-Person Pose Machine\cite{nie2019single} uses Hourglass\cite{newell2016stacked} and PifPaf\cite{kreiss2019pifpaf} uses ResNet\cite{he2016deep}. So, in this paper, we adopt the most commonly used OpenPose model as a baseline to evaluate our proposals.
Finally, Experiments are conducted on the MicroSort COCO dataset. On the COCO minival dataset, the mAP of the OpenPose model by embedding our proposals outperforms the OpenPose baseline model over 5.5\%. Besides, on the COCO test-dev dataset, there is an 11.0\% related accuracy improvement with only 3.5\% related extra computation complexity, which further proves the efficiency of our methods.

The remainder of this paper is organized as follows. We will first explain our second contribution in section 2, because it is the core of the framework presentation. The network architecture we adopted is also given in this part. Section 3 introduces our first contribution, the Self-Supervision based loss function, and Spatial-Sequential Attention-based loss function. The detailed original and modified loss functions are given here. The final loss formula is also given in this part. Section 4 presents the experimental settings, training details as well as the analysis of the results. The discussions of our approach are shown in Section 5.

\section{Resolution Irrelevant Encoding based Bottom-Up Framework}
The baseline system in our experiments is the VGG based multi-stage neural network
with heatmaps and Part Affinity Fields (PAFs), which is called OpenPose\cite{cao2017realtime} and has achieved first place in the COCO keypoints detection task in 2016.  
The architecture of our methods based on the modified baseline system is shown in Fig.  \ref{fig:framework}. The blue block is the content of the baseline system while the red blocks are our modifications. In this section, we will introduce the encoding and decoding parts of the modified baseline model by dividing it into three parts: PAFs, Block inside offsets, and the network architecture.
\subsection{Part Affinity Fields (PAFs)}
The symbol $c$ is adopted for network output channels which is corresponding to joints types (19 in this paper), and $i, j$ are adopted for coordinates of network outputs. The network prediction $F$ contains the Part Affinity Fields $(m,n)$, an offset vector $(x,y)$ and Gaussian heatmaps' score $s$, which is written as
\begin{equation}
F^{c,i,j} = \left \{F_{m}^{c,i,j},F_{n}^{c,i,j},F_{s}^{c,i,j},F_{x}^{c,i,j},F_{y}^{c,i,j}  \right \} .
\end{equation}

\begin{figure}
  \includegraphics[width=\linewidth]{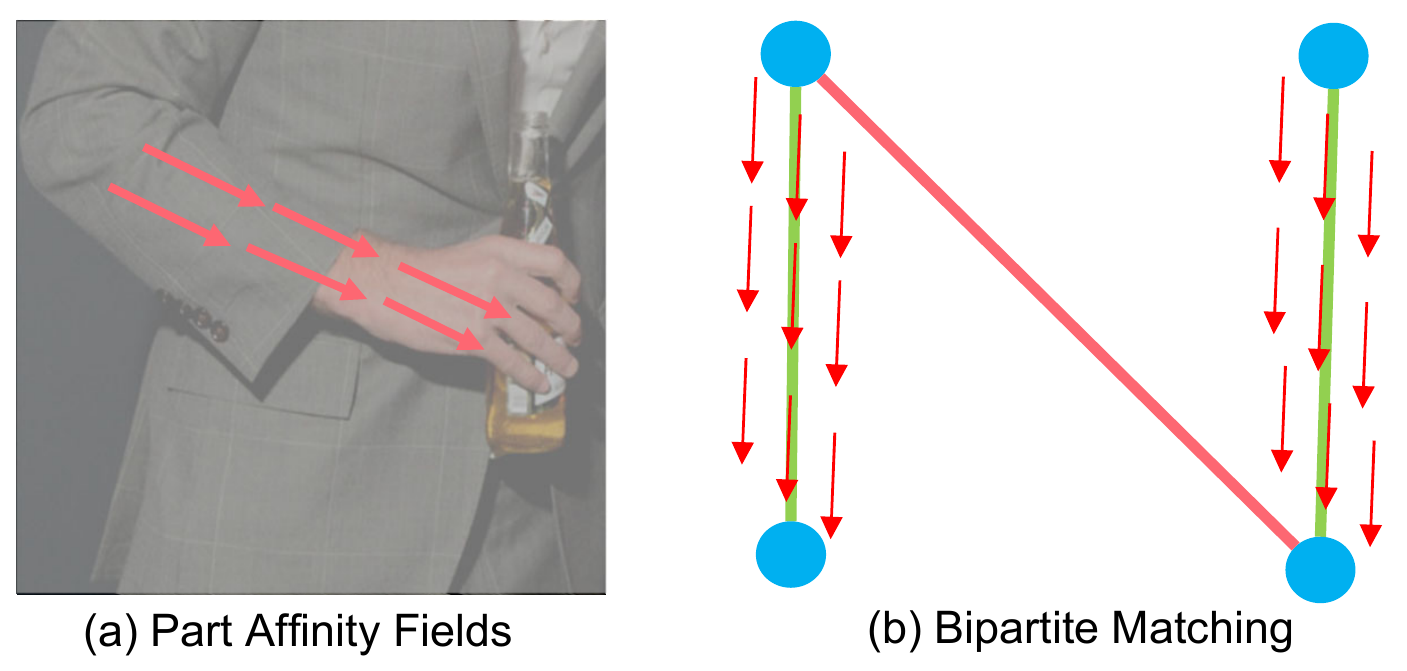}
  \caption{(a) Part Affinity fields (PAFs), the red vectors are unit vectors from elbows to wrists. The (b) shows how does the decoder work on Pafmaps. The red vectors are PAFs, bases on the direction of PAFs, the two green connections are correct and one red connection is incorrect.}
  \label{fig:paf}
\end{figure}

In this subsection, we firstly give the definition of Part Affinity Fields $(m,n)$. As shown in Fig. \ref{fig:paf}, the PAFs are a set of unit vectors that the direction is from one joint to its parent joints, e.g. from one elbow joint to one wrist joint, these two joints belong to one individual in the image. We define the groundtruth of PAFs $(m,n)$ as 
\begin{equation}
F_{m}^{c,i,j}= 
 \begin{cases}
 & \frac{X^{p1} - X^{p2}}{\left \|X^{p1} - X^{p2}  \right \|_{2}} \text{ if pixels on limb }  \\ 
 & 0 \text{ otherwise } 
\end{cases},
\end{equation}
here we adopt $X$ and $Y$ to stand for the real-valued coordinate $x$ and $y$, respectively, $X^{p1}$ and $X^{p2}$ are the parent and child joint for a specific limb. There is a threshold to decide whether a specific pixel is on the limb.

\begin{figure*}
  \includegraphics[width=\linewidth]{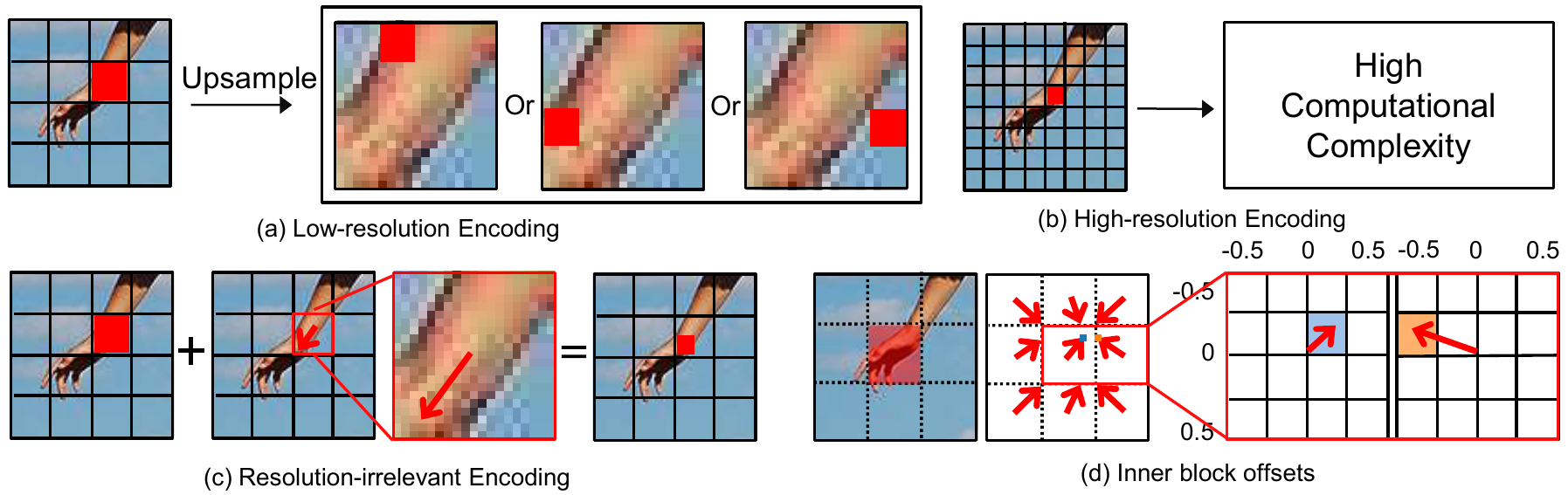}
  \caption{(a) Different upsampling results for Low-resolution encoding. (b) The demerit of high-resolution encoding. (c) Decoding procedure of Resolution-Irrelevant Encoding, conceptual images of the heatmap, our proposed block inside offsets, and the decoding result are shown from left to right. (d) The encoding definition of block inside offsets' groundtruth for a $3\times3$ block; the right two $4\times4$ pixel blocks show the details of offset calculation, the pixel in blue is the groundtruth joint position, and the pixel in orange is the closest pixel to the groundtruth for a neighbor block.}
  \label{fig:framework1}
\end{figure*}

Fig. \ref{fig:paf} (b) shows the processing flow of joints assembling. If we consider a single pair of parts $X^{p1}$ and $X^{p2}$, (e.g., left elbow and left wrist), for a specific limb, finding the optimal association reduces to a maximum weight bipartite graph matching problem. In this graph matching problem, nodes of the graph are the body part detection candidates $X^{c1}$ and $X^{c2}$, and the edges are all possible connections between pairs of detection candidates. Additionally, each edge is weighted by the part affinity aggregate as 
\begin{equation}
S_{k} = \sum_{p}^{P}
\begin{cases}
 & 1 \text{ if direction bias $>$ threshold }  \\ 
 & 0 \text{ otherwise } 
\end{cases}, 
\end{equation}
the $P$ stands for the set of pixels on thfige center line of two potential joints, i.e., the ($X^{p1}$, $X^{c1}$), ($X^{p2}$, $X^{c2}$), ($X^{p1}$, $X^{c2}$) and ($X^{p2}$, $X^{c1}$) are four possible connections of joints $X^{p1}$, $X^{c1}$,$X^{p2}$, $X^{c2}$, the $k$ is index for these possible connections. The direction bias is calculated by 
\begin{equation}
D = |v_{f}||v_{t}|\sin<v_{f},v_{t}>, 
\end{equation}
which is a cross product between network predicted PAFs ($v_{f}$) and the direction vector of two potential joints ($v_{t}$). The threshold is set to 0.5 to keep the same value as OpenPose\cite{cao2017realtime} model. Moreover, the experimental results show the model is robust to this threshold value, the change around 0.2 will not influence the final performance. Finally, matching in a bipartite graph is a subset of the edges chosen in such a way that no two edges share a node. Our goal is to find a matching with maximum weight for the chosen edges. The final $h$ selected connections' $C$ score are 
\begin{equation}
    C_{1},...,C_{h} = max_{h}(S_{1},...,S_{k}).   
\end{equation}
With all limb connection candidates, we can assemble the connections that share the same part detection candidates into full-body poses of multiple people.

\subsection{Block Inside Offsets}

Block inside offsets are proposed to compose a composite encoding field for any Bottom-up pose estimation frameworks. Again, here we summary two main property of the Bottom-up pose estimation frameworks: Adopts heatmaps to estimate joints positions and uses auxiliary predictions to decide the belonging of each joint. Block inside offsets are general to all Bottom-up pose estimation frameworks, which means it can be applied without conceptual level modification, i.e., increasing channels of the network for this new auxiliary prediction. To evaluate all of our proposals simultaneously, we select the OpenPose as the baseline of Bottom-up pose estimation frameworks, the details of the network are listed in the experiment results section.

Besides, the proposed block-inside offsets, as the auxiliary information, will assist the heatmaps to estimate joins position. This composed new framework will not have the network downsampling error in theory, which we called resolution irrelevant encoding (framework) in this paper. The network downsampling error is generated by the difference of the network's input and output images' resolution, where we assume the input images resolution is the same as the original images'.  The extra upsampling procedure is needed because the heatmaps' size is ${8},{16}$ or ${32}$ times smaller than the input images' size, as shown in Fig. \ref{fig:framework1} (a). The theoretical error will be bringing in as these upsampling procedures. There is a 2.5 pixel-distance error on average for the 32× upsampling using the $Intercubic$ interpolation, which is a considerable error in the pose estimation field.

Fig.\ref{fig:framework1} (c) shows the entire processing flow of the joints location. Again, the symbol $c$ is for network output channels which is corresponding to joints types, and symbols $i, j$ are for network output locations. Here we assume the person number in one specific image is $n$, and $P$ stands for pixel coordinates for specific pixels, the detailed pixel value of the prediction heatmap $s$ is 
\begin{equation}
F_{s}^{c,i,j} = \sum_{k=1}^{n}\textup{exp}(\frac{-\left \|P^{c,i,j}-P^{k,c}  \right \|_{2}^{2}}{\sigma^{2}}),
\end{equation}
here $a$ is the closest pixel to the groundtruth joints position in the current block, $f_{d}$ is adopted to stand for the downsampling factor of the network, the calculation formula for each offset value $x$ is  
\begin{equation}
F_{x}^{c,i,j}= 
 \begin{cases}
 & \frac{X^{c,i,j} - X^{a}}{f_{d}} \text{ if } F_{s}^{c,i,j}> 0.4 \\ 
 & 0 \text{ otherwise }  
\end{cases}.
\end{equation}

As in Fig. \ref{fig:framework1} (d), the coarse joints positions are shown in the first $3\times3$ block as the final network outputs. A relative offset will be generated for each pixel around the groundtruth joints position. The fixed threshold 0.4 is set to guarantee only pixels close to the groundtruth joints position are valid by our experiments. Besides, our proposed block inside offsets are different from the other offsets based Bottom-up pose estimation frameworks\cite{papandreou2018personlab,kreiss2019pifpaf}, in our case, the offsets are vectors point to the most likely internal location, which means the required receptive field size will be decreased exponentially as few block's information is enough to estimate the offsets. To increase the convergence speed of the network we set the coordinates $(0,0)$ at the center for each block and assign the output range as $-0.5\sim 0.5$. Furthermore, only the loss in the region with higher heatmaps activation value, instead of full offset-map loss, will be calculated in the training phase, here we also fix this activation value to 0.4.

For the decoder of the RIE, the post-processing algorithm will fix peaks of the heatmap, the final joint positions will be calculated by the corresponding offsets instead of upsampling. Here we assume there are $k$ persons in on an image, the final joints position can be calculated as
\begin{equation}
P_{x\;fixed}^{c,k} = P_{x}^{c,k} + f_{d}\times F_{x}^{c,k}.
\end{equation}

The Fig.\ref{fig:framework1} (c) also shows the decoder of the RIE. If the current pixel is a pixel with maximum activation value, the sub-pixel that its offset vector points to will be the final refined joints position.

\subsection{VGG Based Multi-Stage Network}
\begin{figure}
  \includegraphics[width=\linewidth]{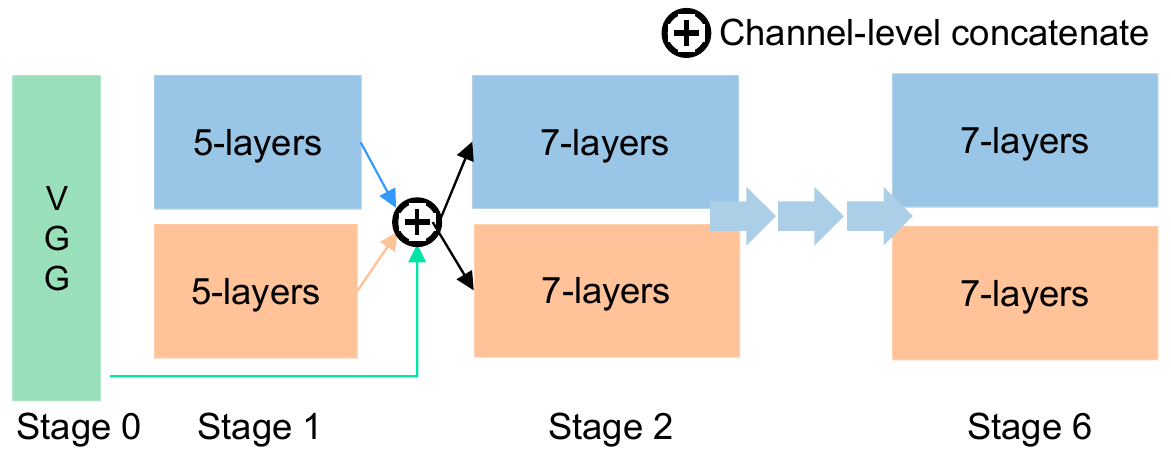}
  \caption{The network architecture of our methods, the layers in each block stands for convolutional layers. The feature maps will be concatenated in channel-level as inputs as the next stage, and there are a total of six stages. }
  \label{fig:vgg}
\end{figure}
The network architecture we adopt is shown in Fig. \ref{fig:vgg}, we keep the same architecture with original OpenPose\cite{cao2017realtime}. The backbone is composed of the first ten layers of the VGG model and two additional fine-tune layers (for reducing the model complexity by cutting channels). For the body part, we assume the set $F_{0}$ is features generated by the backbone network, and $F_{i}^{p}$, $F_{i}^{h}$ stand for the PAFs and heatmaps predictions for the stage $i$, respectively. The outputs of the first stage could be written as
\begin{equation}
    F_{1}^{p}, F_{1}^{h} = \mathcal{F}(F_{0}),
\end{equation}
here, $\mathcal{F}$ stands for the function of the network stage $i$, for the following stage, the outputs could be written as
\begin{equation}
    F_{i}^{p}, F_{i}^{h} = \mathcal{F}(F_{0}\oplus F_{i-1}^{p}\oplus F_{i-1}^{h} ).
\end{equation}
From this formula, the $\oplus$ stands for the concatenate operation at the channel level. The results of the current stage are generated by combining the features of the backbone and the last stage. To realize our block inside offsets, additional 36 channels are added for each $F_{i}^{h}$ in the final layer.

\section{Self-Supervision and spatial-sequential attention based loss}

In this section, we will introduce the training part of our system in detail. The L2 loss will be reorganized by using self-supervised heatmaps and spatial-sequential attention. The original L2 loss of the heatmaps $L_{s}$ in the baseline system could be written as
\begin{equation}
    L_{s} =  \sum_{c}^{C}\sum_{i}^{I}\sum_{j}^{J}(F_{s}^{c,i,j}-T_{s}^{c,i,j})^{2},
\end{equation}
here $C,I,J$ stand for the number of channels, the height and width of images, respectively. The loss ($L_{m},L_{n},L_{x},L_{y}$) for the other predictions ($F_{m},F_{n},F_{x},F_{y}$) are generated by the same formula as $L_{s}$, finally, the original L2 loss $L_{0}$ could be written as
\begin{equation}
    L_{0} = \sum_{r}^{R}L_{r},
\end{equation}
the $R$ is the set of all encoding representatives (i.e., $F_{s},F_{m},F_{n},F_{x},F_{y}$). Different from the Mean Square Error (MSE), we did not calculate the pixel-level average value for $L_{0}$ because the increased factor could be balanced by reducing the learning rate. In the following two subsections, we will introduce two improvements of the $L_{0}$ in detail.

\begin{figure}
  \includegraphics[width=\linewidth]{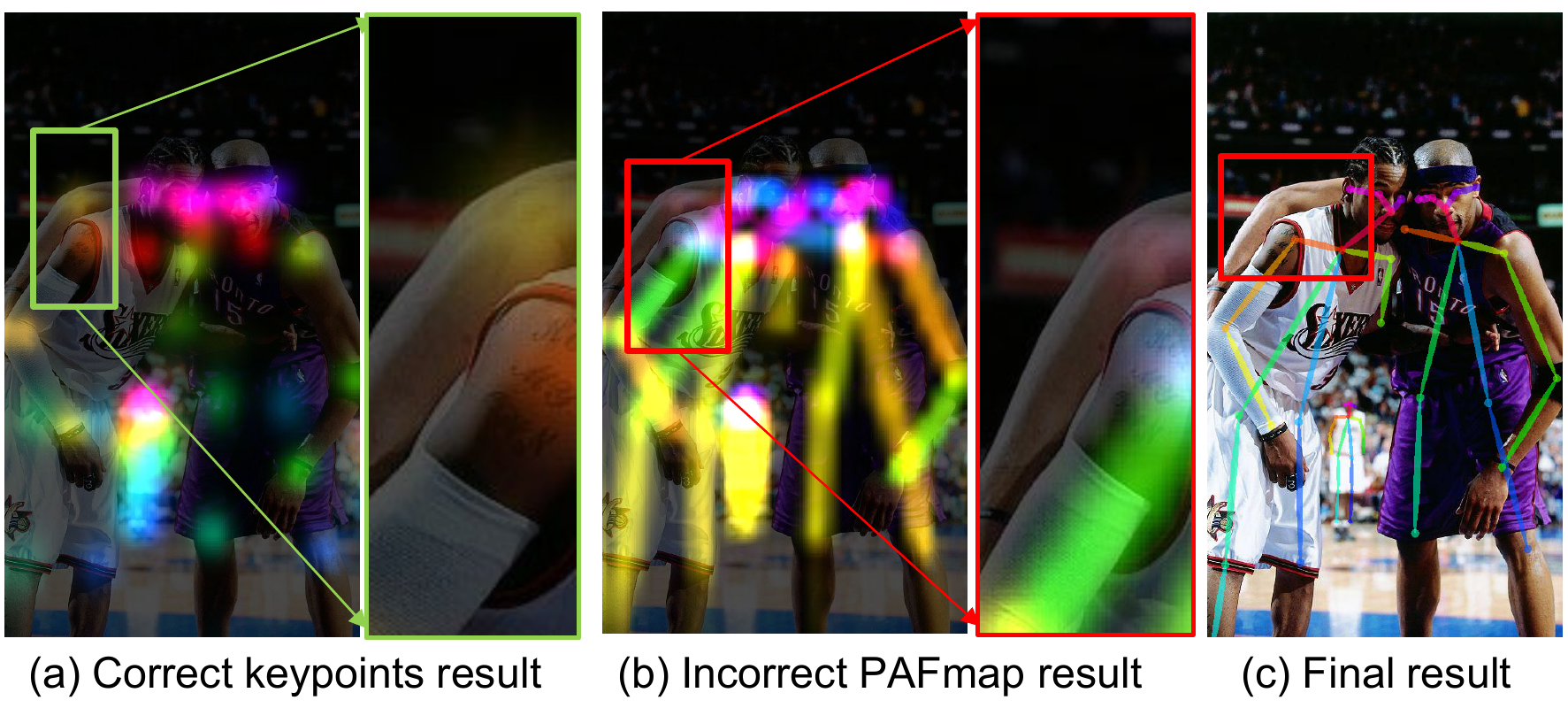}
  \caption{The visualization of failure cases because of contradictions between predictions in one model. The (a) shows the activated region of joints location heatmaps, which is correct for the elbow in (a). However, the activated region of PAFs for the same elbow part is incorrect in (b), which leads the final result (c) to fail.}
  \label{fig:p4demo}
\end{figure}

\begin{figure*}
  \includegraphics[width=\linewidth]{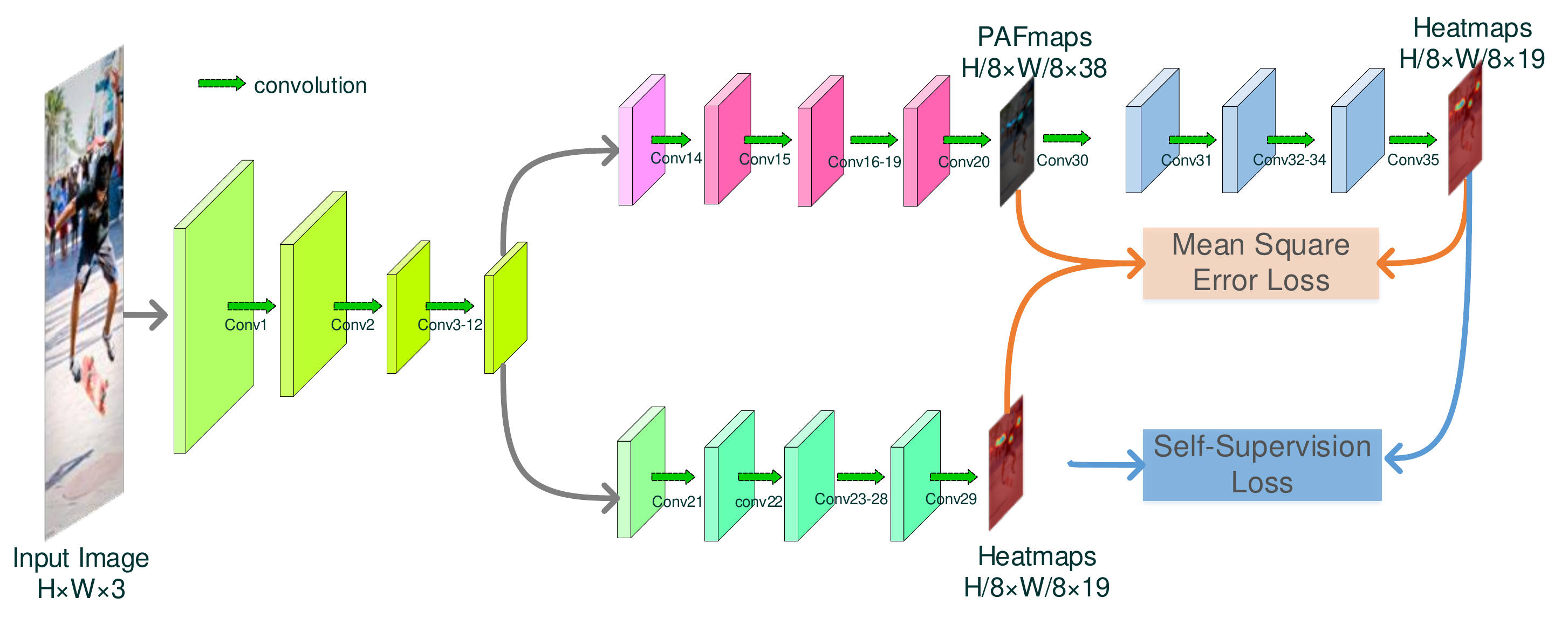}
  \caption{The overall architecture of self-supervision loss. The basic features of input images will be extracted by the backbone, then two independent branches (PAFmap Subnet and Heatmap Subnet) will generate PAFmaps and heatmaps, respectively. The PAFmaps will predict another set of heatmaps for self-evaluating, the self-supervision is realized by KL loss, besides, three L2 loss will also be calculated for the guarantee the accuracy of Heatmaps and PAFmaps.}
  \label{fig:p4}
\end{figure*}

\subsection{Kullback-Leibler Divergence Based Self-Supervision Loss}
The self-supervision loss of our method is composed of self-supervised heatmaps. As shown in Fig. \ref{fig:p4demo}, the circles, and lines of different colors stand for the activated region of different heatmaps and PAFs, respectively. The estimation results of joints are correct but the PAFs results are incorrect, which leads to the final failure. These two predictions are generated by the same features, the only difference is the self-branch that contains 7 convolutional layers, so, there have contradictions between predictions in one model. Instinctively, the activated region of the PAFs endpoints should contain the activated region of heatmaps, which means a correct heatmap could generate a correct PAFmap by connecting associated joints. Besides, the correct PAFmap could generate a correct heatmap by keeping the endpoints region and reducing the redundant region. We test all three combinations of the above-mentioned relationships. The results show using PAFmap to generate heatmap will lead to higher accuracy improvement. A detailed discussion of the different architectural selection will be introduced in the ablation study part. Here we use the most successful architecture to explain the total procedure of our self-supervision loss.

The loss function is based on the architecture shown in Fig. \ref{fig:p4}. Here we ignore the block inside offsets predictions for explaining it more clearly. The PAFmaps predictions will be feed into an additional 3 layers convolutional network and be transferred into corresponding heatmaps. The channel of the PAFmaps predictions will be increased to 128 and in the first two layers and then reduce to 19 in the final layer, which is designed for matching the number of heatmap channels. The main idea of our method is the supervision of generated heatmaps (by the PAFmaps) will lead the extra restriction for the endpoints activated region, which is a kind of explicit supervision to force the network to learn the relationships between two tasks.

However, if we only use L2 loss for the generated heatmaps (by PAFmaps) supervision, the loss will be kept in a high value for the following reasons: In early training epochs, both of PAFmaps and heatmaps are incorrect, which means there are no clear relationships between the incorrect heatmaps and PAFmaps. Force the incorrect PAFmaps to generate correct is a hard and unreasonable task, which will slow the loss convergence and the loss will converge in a high value. If we only use the heatmaps groundtruth to supervise the generated heatmap in later epochs, the target is "how to generate a heatmap more similar to the groundtruth" but it is not equal to enhance the relationship between two tasks. To solve these problems, we use Kullback-Leibler divergence loss to reduce the information distance between network predicted heatmaps and PAFmaps generated heatmaps. As shown in Fig. \ref{fig:p4}, the KL loss could be written as  
\begin{equation}
    L_{kl} = \sum_{c}^{C} \sum_{i}^{I} \sum_{j}^{J} F_{s}^{c,i,j}\log\frac{F_{s}^{c,i,j}}{F_{ps}^{c,i,j}},
\end{equation}
here $F_{ps}$ is the PAFmaps predicted heatmaps, the $L_{kl}$ stands for the distance of predicted distribution $F_{ps}$ to expected distribution $F_{s}$, and based on the knowledge of information theory, we know the $L_{kl}$ is greater than zero and have convex properties, which means it is suitable to be a loss to optimum the networks' weights. The loss function combined $L_{kl}$ and $L_{0}$ is
\begin{equation}
    L_{1} = \gamma L_{s} + L_{ps} + \delta (L_{m}+L_{n}) + L_{kl},
\end{equation}
in this formula, the $L_{ps}$ is the L2 loss to supervise the PAFmaps predicted heatmaps by heatmaps groundtruth, the parameter $\gamma, \delta$ are set for controlling the weight for original loss. To keep the balance of losing weight for PAFmaps and heatmaps, in this paper we always set $\delta=\gamma+1$, then test different $\gamma$ selections' influence to the results, the details will also be introduced in the ablation study part. for the conclusion, we find when $\gamma = 9$ the model will get better performance. In summary, our proposed self-supervision loss is a KL divergence based loss to force the network to learn the relationship between heatmaps and its auxiliary predictions.

\subsection{Spatial-Sequential Attention Loss}
Different from unsupervised attention module\cite{luong2015effective,xu2015show,vaswani2017attention}, our spatial-sequential attention loss is defined by the groundtruth and only be used for the loss calculation, which means: Our attention module is a kind of supervised learning approaches, which could keep higher stability and reliability than the unsupervised approaches. No additional computational complexity is introduced for our attention module, the inference speed will be as same as the original model.

The spatial-sequential attention loss is two-part: Spatial Attention Loss Mask (SALM) and Progressive Direction Distinction (PDD), which are corresponding to the reformations of spatial and sequential loss, respectively. The SALM balance the difficulty between encoding and decoding algorithm, and the PDD adjusts loss weights for the multi-stage loss function.

\begin{figure}
  \includegraphics[width=\linewidth]{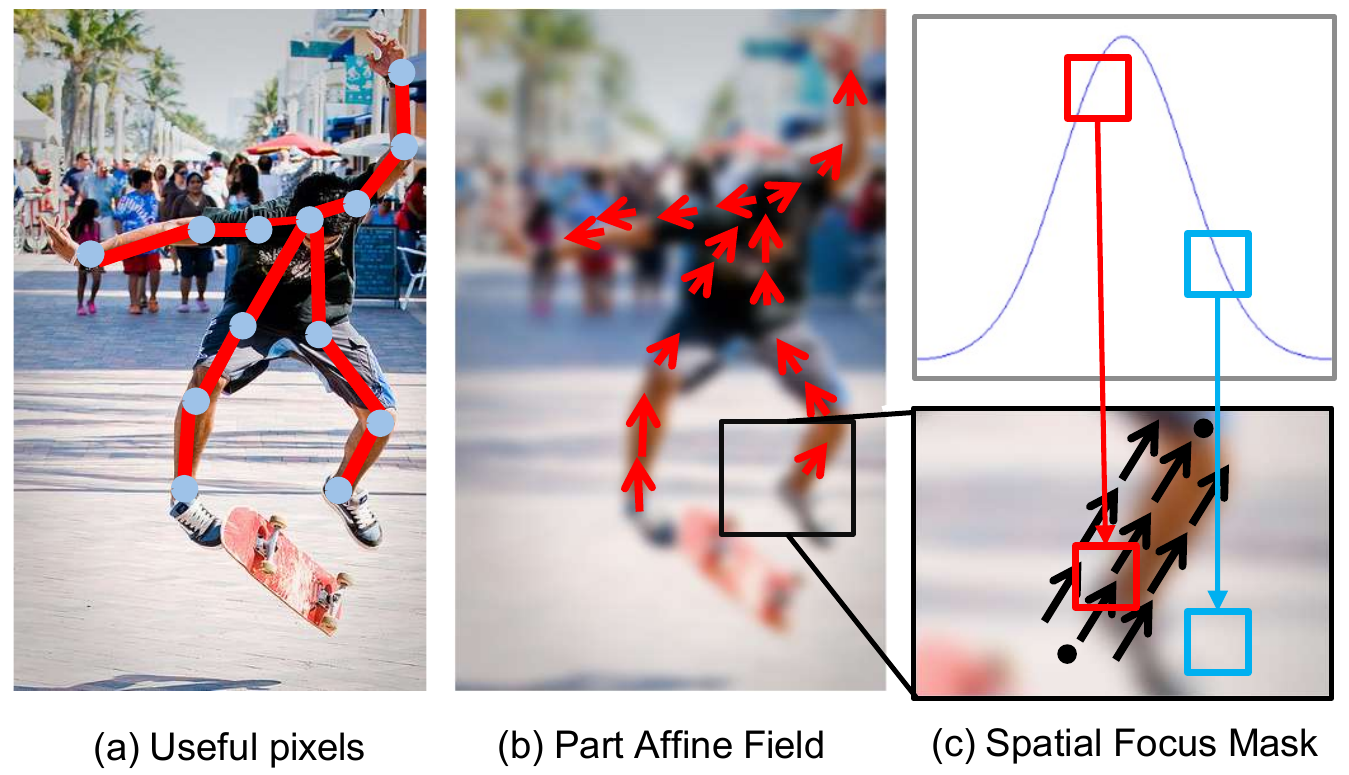}
  \caption{(a) Information utilization ratio unbalance problem, only pixels on the red line will be used for post-processing. The (b), (c) show how does Spatial Attention Loss Mask influence the loss weight during training.}
  \label{fig:p2}
\end{figure}

The proposed spatial-sequential attention loss also has the general applicability for different Bottom-up frameworks. Again, assembling joints group by several auxiliary predictions is one of the main properties for most bottom-up pose estimation frameworks. We also select the OpenPose as the baseline to evaluate our proposals simultaneously. Besides, the Part affinity fields (PAFs), as shown in Fig. \ref{fig:p2} (b), is adopted for analyzing the problem of spatial difficulty unbalance, which is a series of unit direction vectors from parent joints to child's joints.

\textbf{Spatial Attention Loss Mask (SALM)}. The problem of spatial difficulty unbalance is due to the information utilization ratio unbalance between the designation of the encoder and the decoder. Different from heatmaps, for PAFmaps, there are a few pixels used for joints' belonging calculation. However, all pixels in PAFmaps will calculate the loss in the training phase, and a specific scalar (usually 0) should be generated for all negative samples by networks, This kind of task redundancy leads to the inefficient utilization of networks.

Fig. \ref{fig:p2} (a) shows the information utilization region in PAFmaps, the decoder only calculates the connection probability based on pixels between potential connection path, e.g., from one left-wrist to one left-elbow, and the length of the path is only one pixel. Besides, the other negative samples should be regressed as 0. To solve this problem, we rewrite the loss calculation function to a parameterized loss function, the new loss function has a weight for each pixel to explore the best punishment proportion between valuable and worthless information. The modified loss $L_{n}$ for one specific pixel $(c,i.j)$ is calculated as  
\begin{equation}
L_{n}^{c,i,j} = (\alpha\times{exp}(\frac{-\left \| P^{c,i,j}-P^{c,s} \right \|_{2}^{2}}{\sigma ^{2}}) + 1)L_{o}^{c,i,j},
\end{equation}
For Fig. \ref{fig:p2} (c), we marked the projection pixel onto the potential connection line (red vectors) as $s$, the original loss, which is Mean Square Error, is marked by $L_{o}$. The Gaussian distribution based loss weight are set depending on the distance to the connection line to give higher weights for center pixels. Here we take the Gaussian distribution as an example and other methods such as the Ramp function can also be used. We use $\alpha$ and $\sigma$ to control weights between valuable and worthless information. The value of $\sigma$ depends on two factors: the network downsampling factor and the distance between the two connection joints, which is
\begin{equation}
\sigma =\frac{\left \|  P_{parent}^{c,i,j}-P_{child}^{c,i,j}\right \|_{2}}{4f_{d}}.
\end{equation}

For the training phase, a higher weight will be given to the region closed to the potential connection. The more valuable information will be focused for a network by the guidance of SALM supervision.

Furthermore, the first 15 epochs training results show that an additional 1 as bias will increase the model performance instead of setting weights to 0 for all negative regions. the parameter $\alpha$ has been set as one additional untrainable parameter, our loss function will be equal to the original loss when the $\alpha=0$, and the $\alpha=\infty$ means only the loss value for useful information region are focused.

\begin{figure}
  \includegraphics[width=\linewidth]{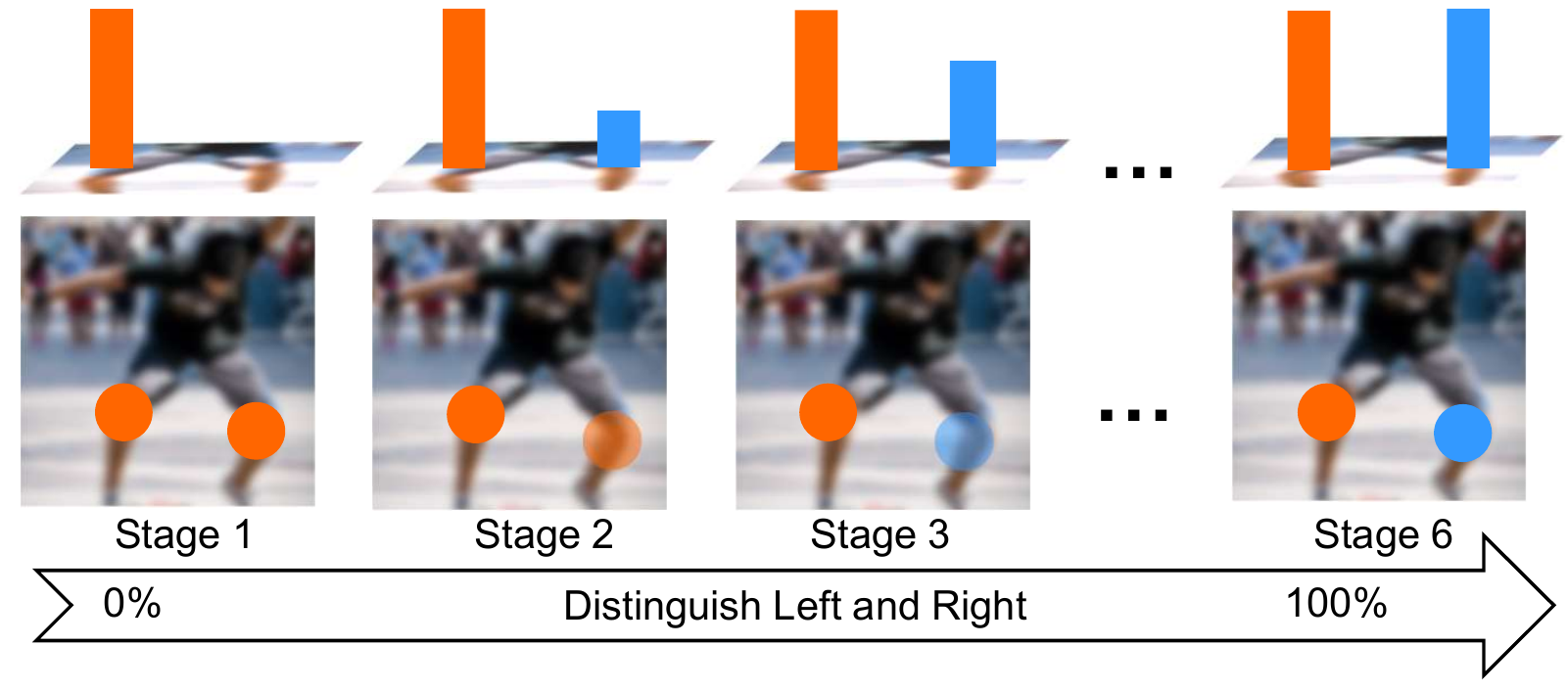}
  \caption{Top: the loss weights for correct and incorrect direction detection results, e.g., the right knee has been detected as the left knee (stage 1), the loss weights are 0. Down: equivalent task representations for PDD.}
  \label{fig:p3}
\end{figure}

\textbf{Progressive Direction Distinction (PDD)}. The same loss functions are usually adopted for Multi-stage networks, which is a type of sequential loss. However, it's more difficult for the first several stages to estimate the precise joints classification results than later stages. For the OpenPose model, there is 0.6\% mAP improvement from stage 3 to 6, which is reported in the \cite{cao2017realtime}. Intuitively, multi-stage networks are more suitable to adopt an easy-to-hard loss design. So, this paper proposes a progressive punishment for incorrect direction detection. e.g., a left knee is estimated as a right knee.

The main idea of PDD is visualized in Fig. \ref{fig:p3}, Detecting joints that ignoring the directions is the equivalent target of the first several stages, which aims to increase the recall of the network. A loss weight from 0\% to 100\% is set for the punishment of the incorrect detection, by assigning higher loss weight, the later stages are supervised to focus on classifying joints' directions. In this paper, all heatmaps, offset-maps and PAFmaps adopt this supervision strategy.

For real loss calculation, the sequence factor $\beta$ is set for changing weights of the progressive punishment for each stage, the new loss function is 
\begin{equation}
L_{p}=\begin{cases}
 & \beta L_{o}\text{ if } direction\;fault\;detection \\ 
 & L_{o}\text{ otherwise } 
\end{cases},
\end{equation}
the $L_{o}$ and $L_{p}$ are our original loss and punishment loss, respectively. The factor $\beta$ increase gradually depending on the distribution function we selected and it contains a series of scalars, e.g., $\left\{0.0,0.2,0.4,...,1.0\right\},\left\{0.06,0.15,0.36,...,1.00\right\}$.

\section{Experiments}
In this section, we will first evaluate the self-supervised heatmaps and spatial-sequential attention module, then conduct the experiments in the original baseline model and our modified baseline model to further prove our methods robustness for different bottom-up pose estimation frameworks.
\subsection{Experiment setting}
\textbf{Dataset.} The MS COCO keypoints challenge dataset\cite{lin2014microsoft} is adopted to evaluate our approaches. There are nearly 5k testing images and 120K training images. Besides, our evaluation metric is the object keypoint similarity (OKS) based mAP\cite{ruggero2017benchmarking}.

\textbf{Network.} For experimental verification, we adopt the OpenPose network as the baseline model, which is consisted of fine-tuned VGG-16\cite{simonyan2014very} and two following 5 or 7 convolutional layers for 6 stages. Additional 36 channels are added for the block inside offsets based resolution irrelevant encoding.

\textbf{Training.} Pytorch\cite{paszke2019pytorch} library is adopt to implement the code. For the training details, the same training and data augmentation hyper-parameters are set as\cite{cao2017realtime} except the GPU number, i.e., the input images will be scaled by the factor from 0.5 to 1.1, then rotate -45 to 45 degrees at random, finally, flip it horizontally with a 50\% probability. The network is trained with mini-batch size 10 on 2 Nvidia GTX 1080Ti GPUs for 5 days. The SGD\cite{johnson2013accelerating} optimizer is adopted with the initial learning 2e-5, which will be decreased depending on current epochs. It will divide 3 after each 170k iterations in a total of 600K iterations. The weight decay is set to 5e-4.

\textbf{Testing.} We adopt multi-scale testing as the testing procedure in \cite{cao2017realtime}. The input image will be padded and warped to 0.5, 1, 1.5, and 2 times of the resolution 368 × 368, 4 prediction maps will be resized and averaged as final network outputs.

\subsection{Ablation Study}
The COCO 2017 validation dataset and COCO 2014 minival subset are used to investigate the effectiveness of our proposed methods, and there are 1.1K images selected randomly from the COCO validation set to compose the COCO 2014 minival subset.

\textbf{Self-Supervision and Spatial-Sequential Attention Loss}. As shown in Table \ref{resp4}. We test three different architecture for the self-supervision loss, i.e., PAFmaps predict heatmaps, heatmaps predict PAFmaps, and both of them. The symbols $L_{p2h}, L_{h2p}, L_{both}$ are stand for these three different architectures, respectively. We keep the same parameter setting ($\gamma = 9$) for all three architectures. The experimental results demonstrate that using heatmaps to predict PAFmaps leads to a decrease in accuracy. Combined with the comparison of the $L_{p2h}$ result, it demonstrates that PAFmaps have more valid information than heatmaps, which lead to PAFmaps predicting heatmaps to become an easier task.

\begin{table}[H]
 \caption{mAP of Self-Supervision Loss}
\label{resp4}
\centering
\begin{tabular}{@{}lccc@{}}
\toprule
\multicolumn{1}{c}{\textbf{Model}} & \textbf{Parameter Selection} & \textbf{Mini2014} & \textbf{Val2017} \\ \midrule
\quad OpenPose       & -    & 58.4              & 57.7             \\
\quad$L_{h2p}$     & $\gamma = 9$    & 56.6              & 56.0             \\
\quad$L_{both}$    & $\gamma = 9$    & 58.7              & 57.9             \\
\quad$L_{p2h}$     & $\gamma = 9$    & \textbf{59.2}     & \textbf{58.6}    \\
\quad$L_{p2h}$     & $\gamma = 0$    & 58.4              & 57.6             \\
\quad$L_{p2h}$     & $\gamma = 99$   & 59.0              & 58.5             \\ \bottomrule
\end{tabular}
\end{table}

Next, we evaluate the results for different parameter selection. As the result shown in Table \ref{resp4}, when the weights of original L2 loss and $L_{kl}$ are nearly the same ($\gamma=1$), the performance almost has not been improved. The result also shows the best choice of the value of $\gamma$ is between $1~100$.

For spatial-sequential attention loss, the evaluation also is divided into two parts: Spatial Attention Loss Mask (SALM) and Progressive Direction Distinction (PDD).

For SALM, different parameter $\alpha$ settings experiments are conducted. Table \ref{my-labelSALM} shows two COCO subset's results: The 1-100 is the range of best setting for $\alpha$, and our method outperforms the OpenPose baseline by 3.0\% when $\alpha=10$ in two datasets. Besides, we get the result lower punishment will not gain more benefits for the network from the $\alpha=100$ results.

\begin{table}[H]
\caption{mAP of Spatial-Sequential Attention Loss}
\label{my-labelSALM}
\centering
\begin{tabular}{@{}lcccc@{}}
\toprule
\multicolumn{1}{c}{\textbf{Model}} & \textbf{SALM} & \textbf{PDD}& \textbf{Mini 2014} & \textbf{Val 2017} \\ \midrule
\quad OpenPose       & $\mathrm{\alpha=0}$ &Uniform           & 58.4               & 57.7              \\
\quad $\mathrm{Ours_{\alpha=1}}$           & $\mathrm{\alpha=1}$ &Uniform           & 58.7               & 57.9              \\
\quad $\mathrm{Ours_{\alpha=10}}$           & $\mathrm{\alpha=10}$&Uniform          & \textbf{61.4}      & \textbf{60.5}     \\
\quad $\mathrm{Ours_{\alpha=100}}$           & $\mathrm{\alpha=100}$&Uniform         & 60.8               & 59.9              \\
\quad $\mathrm{Ours_{Linear}}$           & $\mathrm{\alpha=0}$&Linear            & 59.5               & 58.7              \\
\quad $\mathrm{Ours_{Quadric_{a}}}$           & $\mathrm{\alpha=0}$&$\mathrm{Quadric_{a}}$               & 58.9               & 58.2              \\
\quad $\mathrm{Ours_{Quadric_{b}}}$           & $\mathrm{\alpha=0}$&$\mathrm{Quadric_{b}}$               & \textbf{59.7}      & \textbf{58.8}   \\
\bottomrule
\end{tabular}
\end{table}

For PDD, we adopt different $\beta$ to generate different progressive punishment functions, and there is six stages as the same as the baseline model. Table \ref{my-labelSALM} shows results, and $\left \{ 0.0,0.2,0.4,0.6,0.8,1.0 \right \}$ is the corresponding $\beta$ linear setting, the $\left \{ 0.00,0.45,0.69,0.85,0.95,1.00\right \}$ and quadratic-b is $\left \{ 0.00,0.05,0.15,0.31,0.65,1.00\right \}$ are quadratic-a and quadratic-b, respectively. We found the quadratic-b are most efficient from results in minival 2014, which has 1.3\% improvement. Furthermore, the results shows the lesser punishment in early stages of the multi-stage network will preform better.

\textbf{Block Inside Offsets Based Resolution Irrelevant Encoding}. We first evaluate the validity of block inside offset by comparing results with and without block inside offsets based resolution irrelevant encoding on the OpenPose model. Table \ref{my-label1} shows results with different network downsampling factors on two COCO subset, there are 1.5\% and 1.4\% mAP improvement in 8× case for our proposed Resolution Irrelevant Encoding, and when the downsampling factor is 16, there is more than 3.0\% mAP improvement. The reason is the theoretical error of 16× upsampling is larger than 8×. In a summary, our proposed block inside offsets based resolution irrelevant encoding can better guide the network to estimate the precise joint's position for high downsampling factor networks.

Besides, we evaluate our proposed loss functions' robustness by embedding self-supervision and spatial-sequential attention loss in the original OpenPose model and our modification model. The results is also shown in the Table \ref{my-label1}, the $\mathrm{OpenPose_{w}}, \mathrm{Ours_{w}}$ means baseline and our model with self-supervision and spatial-sequential attention loss. The results show our proposed loss function could improve the model performance for different encoding methods, which means it could be embedded into different Bottom-Up based pose estimation approaches.

\begin{table}[H]
 \caption{mAP of Resolution Irrelevant Encoding}
\label{my-label1}
\centering
\begin{tabular}{@{}lccc@{}}
\toprule
\multicolumn{1}{c}{\textbf{Model}}    & \textbf{Downsample Factor} & \textbf{Mini 2014} & \textbf{Val 2017} \\ \midrule
  \quad OpenPose & $f_{d}=8$                        & 58.4               & 57.7              \\
\quad $\mathrm{Ours_{f_{d}=8}}$     & $f_{d}=8$                        & \textbf{59.9}      & \textbf{59.1}     \\
\quad OpenPose          & $f_{d}=16$                       & 44.6               & 43.9              \\
\quad $\mathrm {Ours_{f_{d}=16}}$     & $f_{d}=16$                       & \textbf{47.8}               & \textbf{46.9}  \\
\quad $\mathrm{OpenPose_{w}}$     & $f_{d}=8$                       & \textbf{63.1}               & \textbf{62.5} \\
\quad $\mathrm{Ours_{w}}$     & $f_{d}=8$                       & \textbf{63.9}               & \textbf{62.8} \\ \bottomrule
\end{tabular}
\end{table}

\subsection{Results in COCO}
\textbf{Accuracy.} In this part we also keep the same evaluation standard as\cite{cao2017realtime}, i.e., the COCO minival 2014 and test-dev dataset are selected for the final evaluation. Table \ref{my-label} shows the final results. Our single model reaches improvement of mAP by 5.5\% and outperforms refinement results by the convolutional pose machine\cite{wei2016convolutional} over 2.8\% in the minival 2014, which is better than our previous work\cite{liu2020resolution} by 0.6\%.

\begin{table}[]
 \caption{Final Performance on COCO Dataset }
\label{my-label}
\centering
\begin{tabular}{llllll}
\toprule
\multicolumn{1}{c}{\textbf{Model}}                  & \multicolumn{1}{l}{\textbf{AP}}   & $\mathrm{AP^{50}}$       & $\mathrm{AP^{75}}$  & $\mathrm{AP^{M}}$          & $\mathrm{AP^{L}}$        \\ \midrule
\multicolumn{6}{c}{COCO minival 2014}                                                                                                                 \\
\multicolumn{1}{l}{OpenPose}                & \multicolumn{1}{l}{58.4}          & 81.5          & 62.6          & 54.4          & 65.1          \\
\multicolumn{1}{l}{+ CPM refine}        & \multicolumn{1}{l}{61.0}          & 84.9          & 67.5          & 56.3          & \textbf{69.3} \\
\multicolumn{1}{l}{Ours}                    & \multicolumn{1}{l}{\textbf{63.9}} & \textbf{86.8} & \textbf{70.6} & \textbf{63.2} & 68.1          \\ 
\multicolumn{6}{c}{COCO test-dev}                                                                                                                 \\
\multicolumn{1}{l}{OpenPose}                & \multicolumn{1}{l}{56.6}          & 80.3          & 61.0          & 53.9          & 64.7          \\
\multicolumn{1}{l}{+ refine\&embedding} & \multicolumn{1}{l}{61.8}          & 84.9          & 67.5          & 57.1          & \textbf{68.2} \\
\multicolumn{1}{l}{Ours}                    & \multicolumn{1}{l}{\textbf{62.8}} & \textbf{86.3} & \textbf{70.2} & \textbf{61.8} & 67.9          \\ \bottomrule
\end{tabular}
\end{table}

In the test-dev dataset, there is more than 6.2\% mAP improvement compare to our single model performance to the official release model in OpenPose GitHub project\cite{cao2017realtime}. our single model performance still outperforms over 1.0\% for the result with CPM refinement and model embedding.

\begin{figure}[H]
  \includegraphics[width=\linewidth]{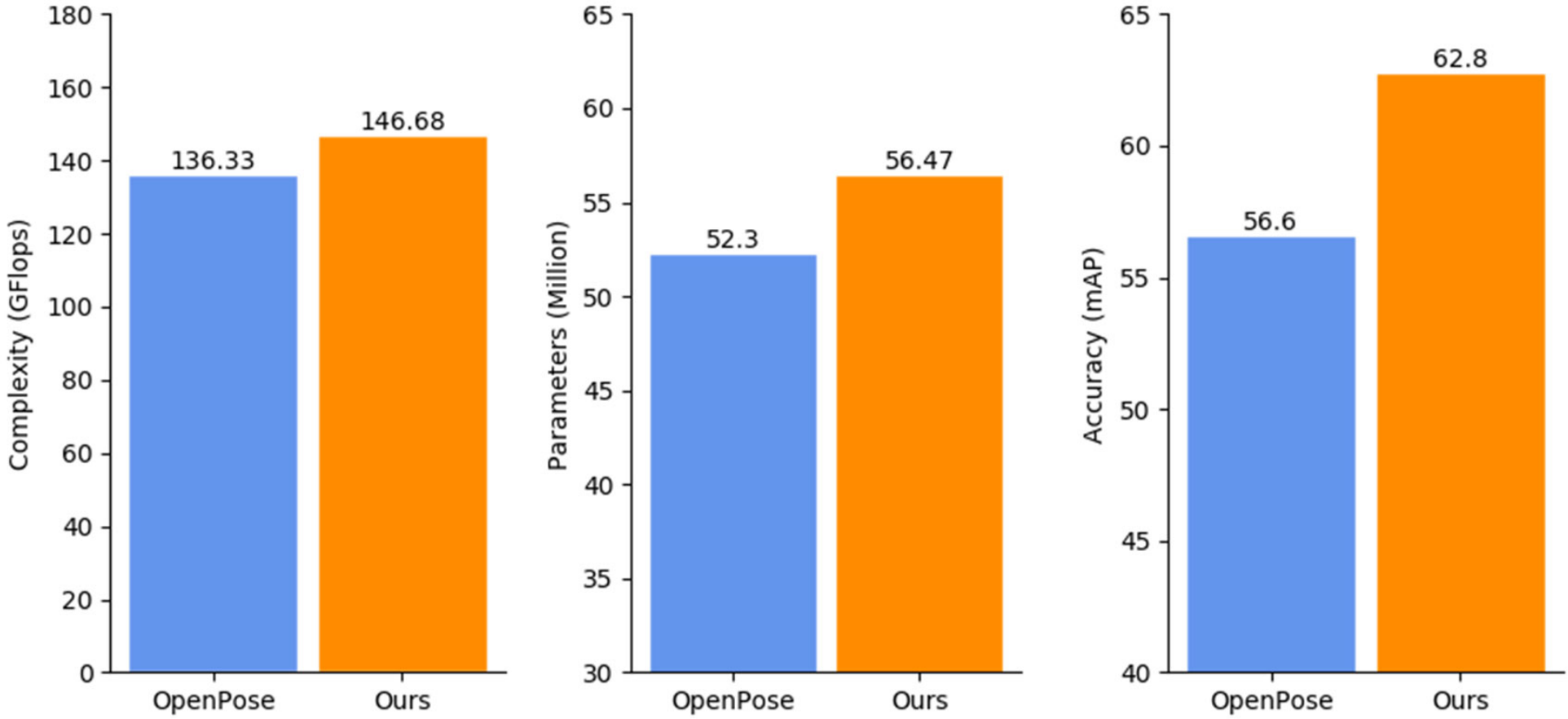}
  \caption{Statistics of GFLOPs, parameter numbers and accuracy for the baseline and baseline training with our methods.}
  \label{fig:com}
\end{figure}

\textbf{Complexity.} Fig. \ref{fig:com} visualizes the computation complexity of the model, number of parameters, and testing accuracy between the OpenPose model and the model training with our proposals. The computation complexity in the training phase will be increased by spatial-sequential attention loss, but the spatial-sequential attention loss leads to a zero increase in computation resources in the inference phase. Additional 36 channels are added to the final layers of each stage for the block inside offsets-based resolution irrelevant encoding, which lead to computational complexity and parameter number increase by 3.5\% and 4.4\%, respectively. Overall, compare with \cite{xiao2018simple} that increases 47.5\% complexity with 1.4\% accuracy improvement, our methods have 11.0\%  accuracy improvement, which is more computationally efficient.

\begin{figure}[]
  \includegraphics[width=\linewidth]{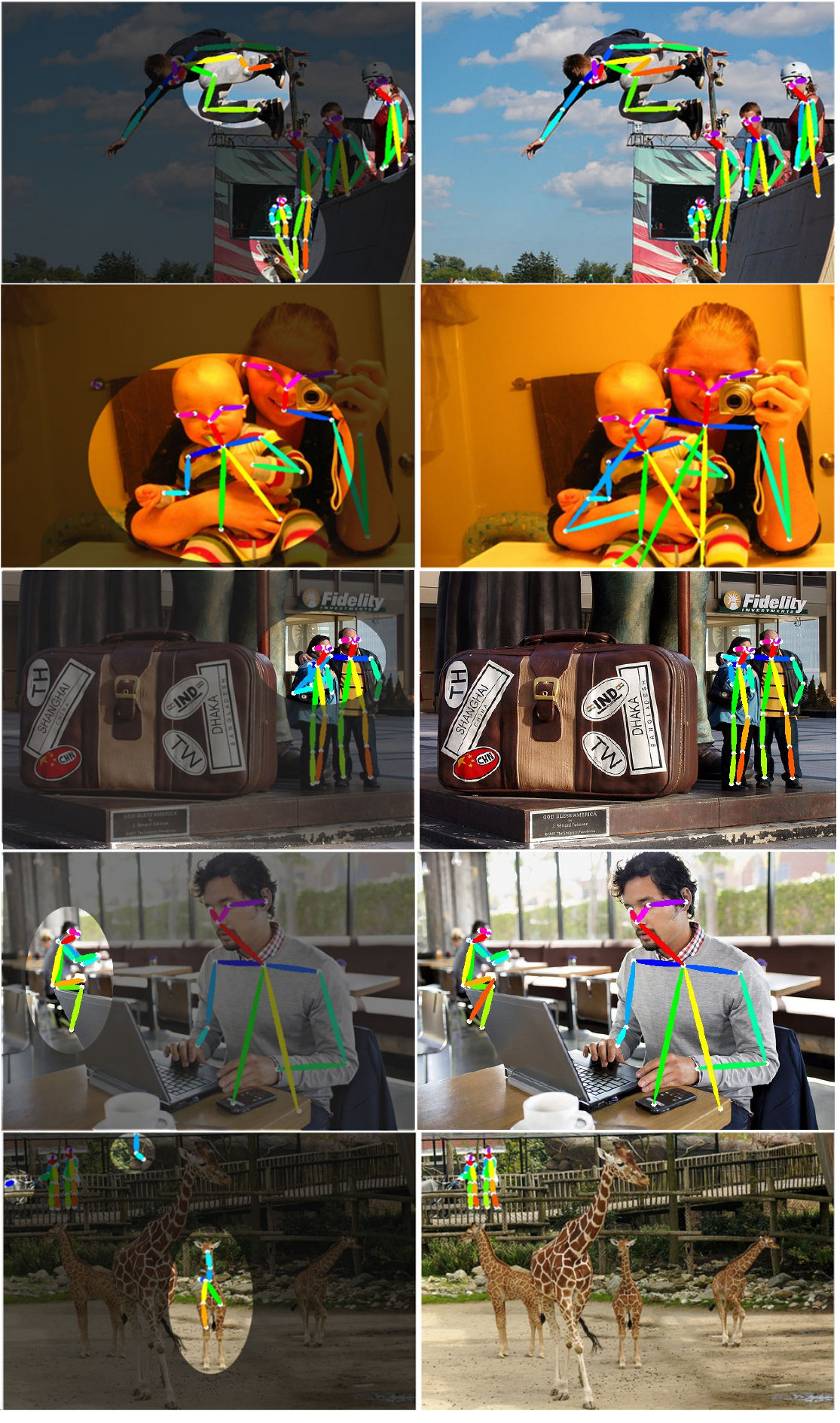}
  \caption{Comparison of OpenPose (Left) and OpenPose trained with our methods (Right) in typical challenge scenes: abnormal pose; occlusion and overlap; background error.}
  \label{fig:res_CCC}
\end{figure}

\begin{figure*}
  \includegraphics[width=\linewidth]{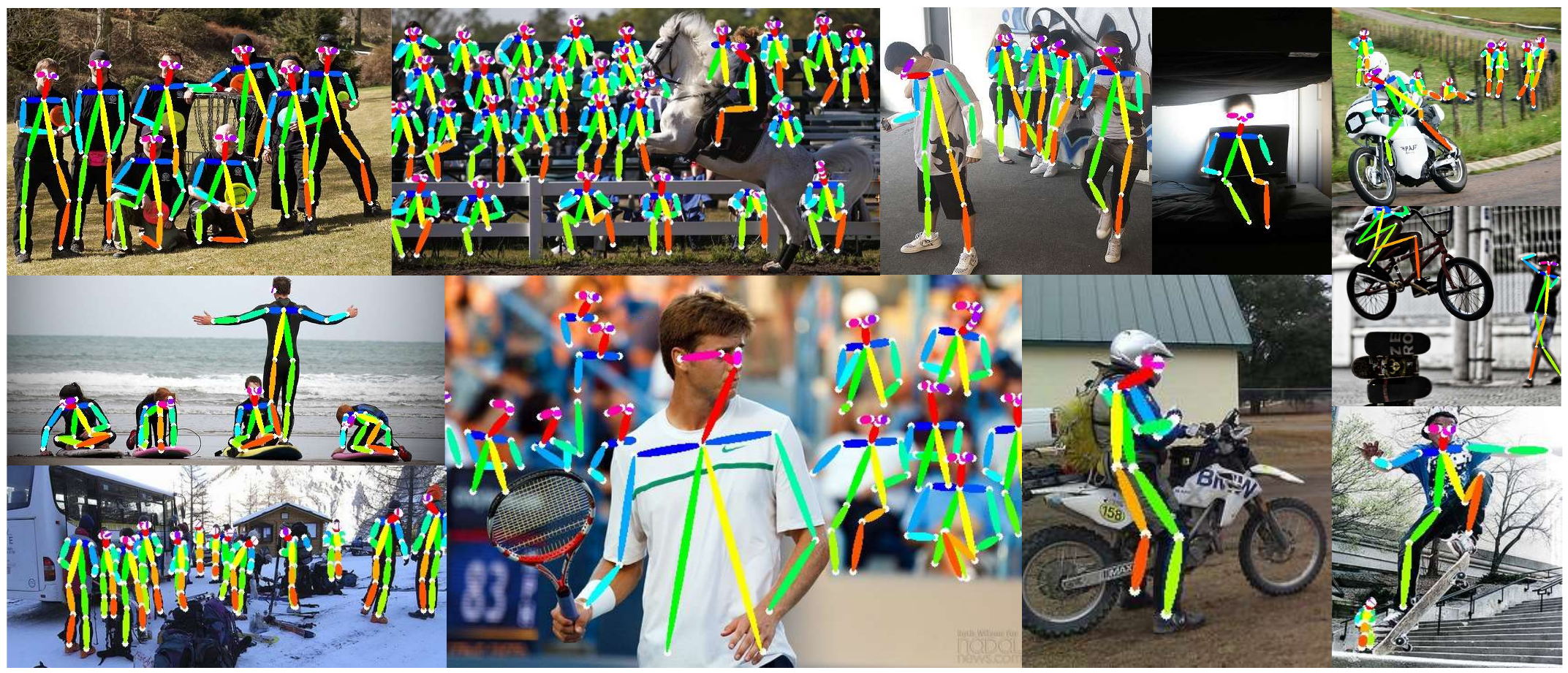}
  \caption{Visualization of our results on the COCO dataset}
  \label{fig:rescoco}
\end{figure*}

\textbf{Visual Analysis.} In Fig. \ref{fig:res_CCC}, the baseline and our model's results are visualized. Results demonstrate our network-agnostic supervision improves the performance not only in some general scenes but also in several typical challenge scenes such as background errors, occlusion, and abnormal pose. Fig. \ref{fig:rescoco} shows more results on the COCO dataset generated by our methods.    

\section{Conclusion}
In this paper, we propose two supervision strategies for high-performance human pose estimation belonging to two different supervision levels, which is based on Self-Supervision and Spatial-Sequential Attention Loss and Block inside offset based Resolution Irrelevant Encoding.
The self-supervision and spatial-sequential attention based loss function is proposed to solve the in lower-level supervision (low features utilization for images and network stages, contradictions between predictions). A new combination of predictions composed of heatmaps, Part Affinity Fields (PAFs), and block-inside offsets is also proposed to fix pixel-level joints positions. Finally, the experiment results demonstrate our methods will boost the model performance with a little extra computation cost.

Besides, the proposed supervision strategies are network-agnostic, which is able to be applied to different Bottom-Up pose estimation frameworks. In this paper, we adopt a unity framework (OpenPose) to demonstrate their effectiveness simultaneously. For more details, the Resolution Irrelevant Encoding is able to be adopted for all pose estimation approaches using the heatmaps to locate joints, and the Self-Supervision and Spatial-Sequential Attention Loss can be adopted for other one-stage frameworks such as Ike2D\cite{luo2019end}. It is suitable for other networks such as Hourglass\cite{newell2016stacked}. Our future work will extend our supervision strategies in the state-of-the-art Bottom-Up architecture to reach higher final performance in this field. 

\section*{Acknowledgment}

This work was jointly supported by the Waseda University Grant for Special Research Projects under Grants 2020C-657 and 2020R-040, and the National Natural Science Foundation of China under grant 62001110.

\ifCLASSOPTIONcaptionsoff
  \newpage
\fi

\bibliographystyle{./bio/IEEEtran}
\bibliography{ref}
\newpage
\begin{IEEEbiography}[{\includegraphics[width=1in,height=1.25in,clip,keepaspectratio]{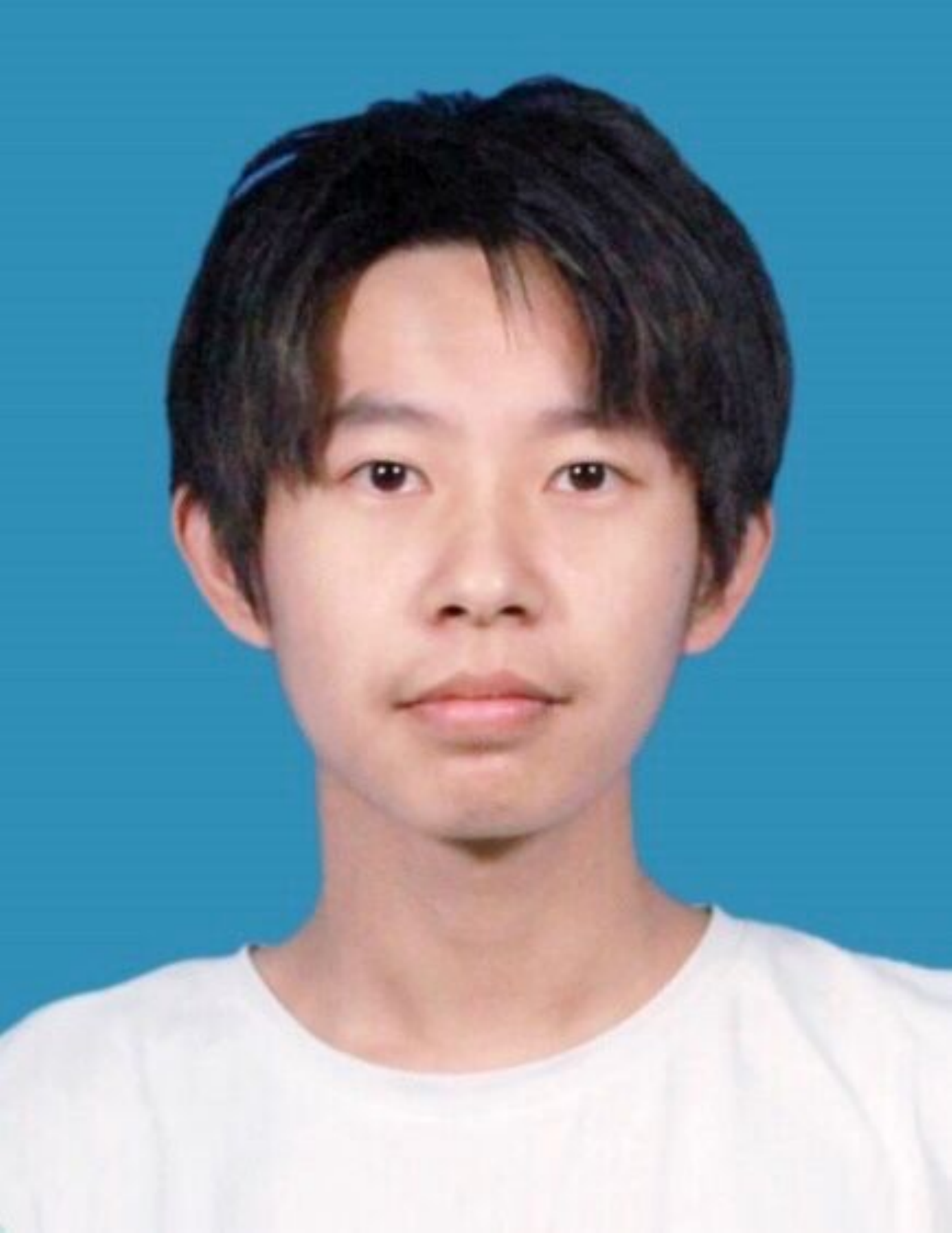}}]{Haiyang Liu} is a Ph.D. student in information science and technology at The University of Tokyo, Japan. He received his M.E. degree from the Graduate School of Information, Production
and Systems, Waseda University, Japan, in 2020, and the B.E. degree in Instrument Science and Engineering from Southeast University, China, in 2019. His research interests are mainly in computer vision, image \& audio signal processing and deep learning.    
\vspace{-10 mm}

\end{IEEEbiography}
\begin{IEEEbiography}[{\includegraphics[width=1in,height=1.25in,clip,keepaspectratio]{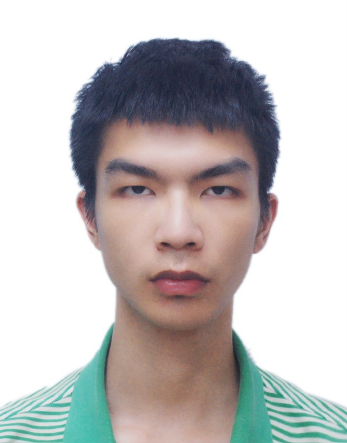}}]{Dingli Luo}
received the B.E. degree in software engineering from Electronic Science and Technology of China, China, in 2017, and the M.E. degree from Graduate School of Information Production and Systems, Waseda University, Japan, in 2020. His research focuses on computer vision with deep learning and computer graphics. 
\end{IEEEbiography}
\vspace{-10 mm}

\begin{IEEEbiography}[{\includegraphics[width=1in,height=1.25in,clip,keepaspectratio]{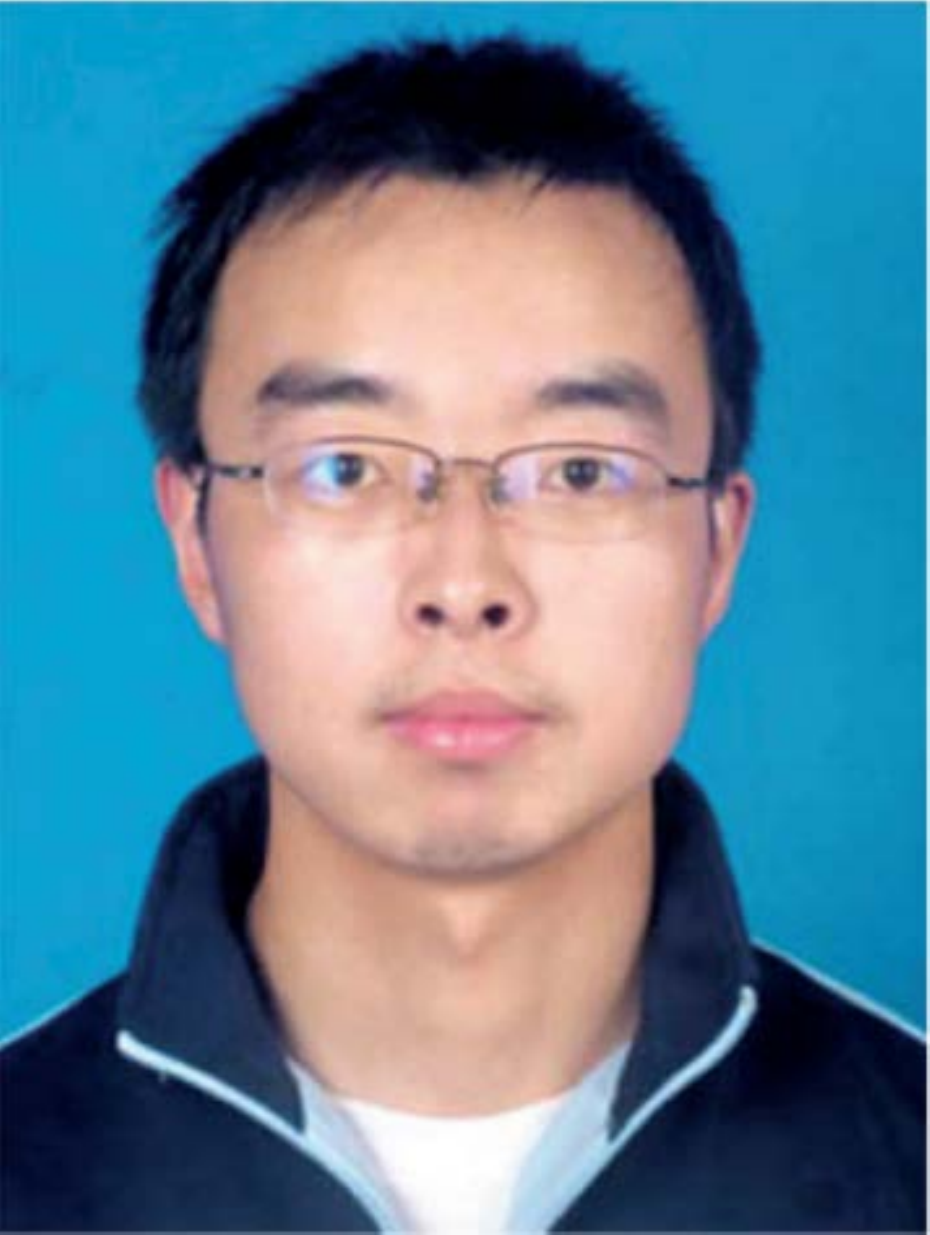}}]{Songlin Du}
received the Ph.D. degree from the Graduate School of Information, Production
and Systems, Waseda University, Kitakyushu,Japan. He is currently with the School of Automation, Southeast University, Nanjing, China. His research interests include visual feature representation and related hardware implementation. He received the Best Paper Award at ISPACS2017. He is a member of the IEEE.
\end{IEEEbiography}
\vspace{-10 mm}

\begin{IEEEbiography}[{\includegraphics[width=1in,height=1.25in,clip,keepaspectratio]{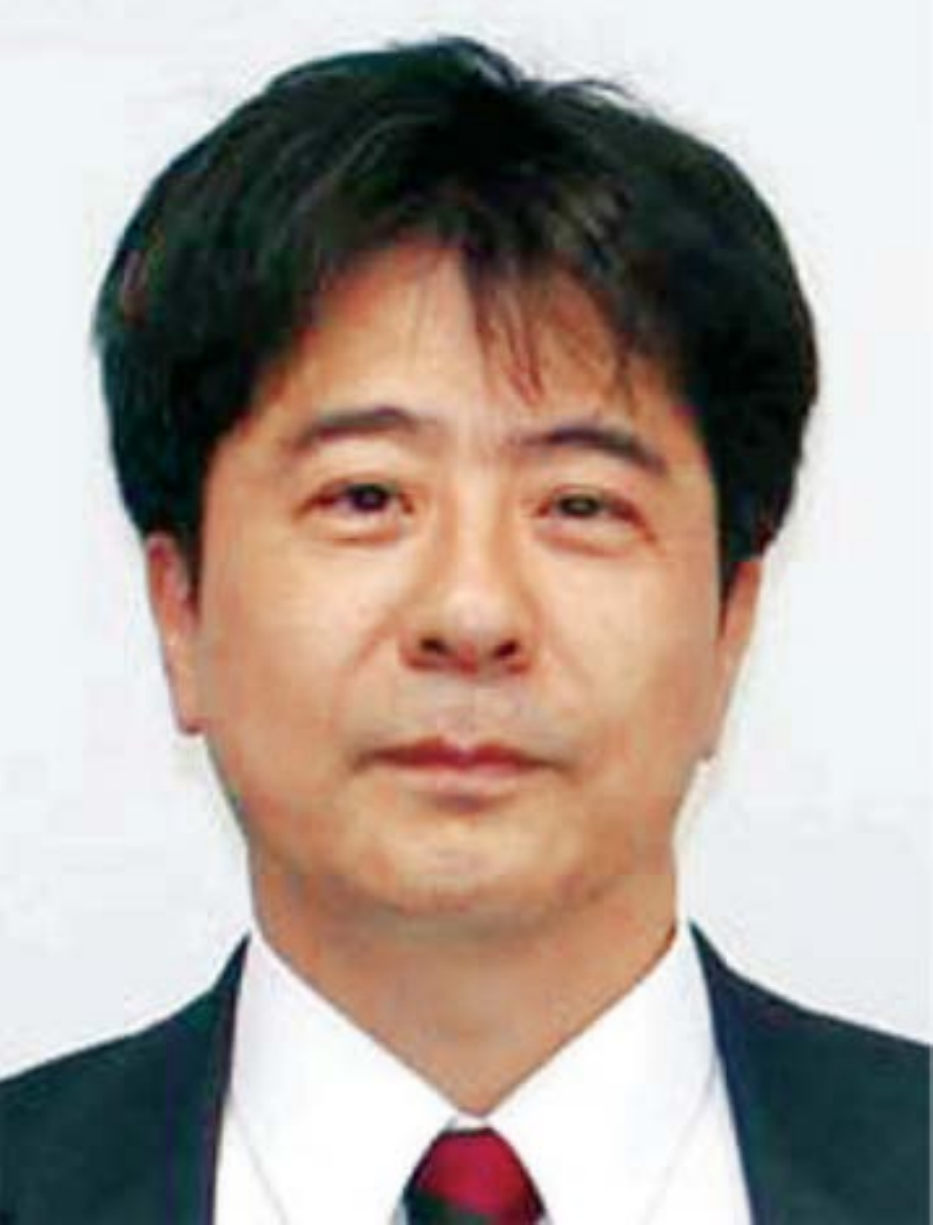}}]{Taseshi Ikenaga}
received his B.E. and M.E. degrees in electrical engineering and Ph.D. degree in information \& computer science from Waseda University, Tokyo, Japan, in 1988, 1990, and 2002, respectively. He joined LSI Laboratories, Nippon Telegraph and Telephone Corporation (NTT) in 1990, where he had been undertaking research on the design and test methodologies for high performance ASICs, a real-time MPEG2 encoder chip set, and a highly parallel LSI \& system design for image understanding processing. He is presently a professor in the system integration field of the Graduate School of Information, Production and Systems, Waseda University. His current interests are image and video processing systems, which covers video compression (e.g. H.265/HEVC, SHVC, SCC), video filter (e.g. super resolution, noise reduction, high-dynamic range imaging), and video recognition (e.g. sport analysis, feature point detection, object tracking). He is a senior member of the Institute of Electrical and Electronics Engineers (IEEE), a member of the Institute of Electronics, Information and Communication Engineers of Japan (IEICE) and the Information Processing Society of Japan (IPSJ).
\end{IEEEbiography}

\end{document}